\documentclass[preprint,12pt]{elsarticle}

\usepackage{amssymb}
\usepackage{kpfonts}
\usepackage{booktabs}
\usepackage{graphicx}
\usepackage{subcaption}

\begin{document}

\begin{frontmatter}

\title{A Granular Grassmannian Clustering Framework via the Schubert Variety of Best Fit}

\author[inst1]{Karim Salta}
\affiliation[inst1]{organization={Colorado State University, Department of Mathematics},
            addressline={Louis R. Weber Building
1874 Campus Delivery},
            city={Fort Collins},
            postcode={80523}, 
            state={Colorado},
            country={USA}, 
            mail={, kkarimov@gmail.com}}

\author[inst2]{Michael Kirby}
\affiliation[inst2]{organization={Colorado State University, Department of Mathematics},
            addressline={Louis R. Weber Building
1874 Campus Delivery},
            city={Fort Collins},
            postcode={80523}, 
            state={Colorado},
            country={USA}, 
            mail={, michael.kirby@colostate.edu}}
\author[inst3]{Chris Peterson}
\affiliation[inst3]{organization={Colorado State University, Department of Mathematics},
            addressline={Louis R. Weber Building
1874 Campus Delivery},
            city={Fort Collins},
            postcode={80523}, 
            state={Colorado},
            country={USA}, 
            mail={, peterson@math.colostate.edu}}

\begin{abstract}
In many classification and clustering tasks it is useful to compute a geometric representative for a dataset or a cluster, such as a mean or median. When datasets are represented by subspaces, these representatives become points on the Grassmann or flag manifold, with distances induced by their geometry, often via principal angles.  

We introduce a subspace clustering algorithm that replaces subspace means with a trainable prototype defined as a \textit{Schubert Variety of Best Fit} (SVBF) — a subspace that comes as close as possible to intersecting each cluster member in at least one fixed direction. Integrated in the Linde–Buzo–Grey (LBG) pipeline, this SVBF-LBG scheme yields improved cluster purity on synthetic, image, spectral, and video action data, while retaining the mathematical structure required for downstream analysis.
\end{abstract}

\tnotetext[LBG]{Linde-Buzo-Grey clustering algorithm}
\tnotetext[SVBF]{Schubert Variety of Best Fit}


\end{frontmatter}

\section{Introduction}
\label{sec:intro}

There has been sustained interest in leveraging the geometry of subspace manifolds to analyze collections of datasets \cite{hamm2008grassmann,BeDrChKiKlPe,huang2015projection,motionrecognition,srivastava2004bayesian,srivastava2004bayesian,NeighbPresIm}. The key feature shared by such techniques is that data is encoded by orthonormal bases, with distances defined by the manifold geometry — Grassmannian methods, for example, rely on principal angles between subspaces, while related frameworks appear in illumination-invariant image analysis, subspace tracking, graph embeddings, and set recognition.

Vector quantization methods such as LBG clustering \cite{LBG} iteratively assign samples to the closest cluster prototype, then update the prototype based on the structure of its assigned points. Prior extensions of LBG to the Grassmannian compute cluster centers as subspace means or medians (e.g., flag mean \cite{stiverson2019subspace} and flag median \cite{mankovich2022flag}).  

Instead of classical centroids, our model updates each cluster prototype to be the solution of the SVBF intersection optimization problem \cite{ourfirst}. The resulting algorithm maintains the tone and structure of existing manifold clustering pipelines while offering a more geometrically flexible model of a cluster representative.

\section{Related work}
\label{sec:relatedwork}
Existing K-means variants on the Grassmann manifold fall broadly into kernelbased and direct-optimization categories. The most competitive direct methods include intrinsic optimization on the manifold itself, extrinsic embedding
into ambient Euclidean space, or hybrid geodesic updates.
Several works propose different Grassmannian means such as the Karcher
mean (l2 minimizer of principal angle norms) \cite{geommean}, the chordal-distance flag mean
(closed-form via SVD) \cite{draper2014flag}, and iterative IRLS-weighted flag median (current stateof-the-art for subspace LBG clustering) \cite{mankovich2022flag}.
Our method is also a direct, geometry-driven optimization, but differs in
that it constructs an adaptive prototype optimizing subspace incidence conditions rather than minimizing a classical Fréchet-type mean.

\section{Background}
\label{sec:background}

\subsection{Principal Angles}
Let $U$ and $V$ be subspaces of $\mathbb R^n$ with orthonormal basis matrices $A$ and $B$ respectively. If $\sigma_1\geq\dots\geq\sigma_r$ are the singular values of $A^TB$, then $\sigma_i=\cos\Theta_i$ are the cosines of the principal angles $\Theta_i$.  
We write $\Theta(U,V)=[\Theta_1,\dots,\Theta_r]$ and $\sin\Theta(U,V)=[\sin\Theta_1,\dots,\sin\Theta_r]$.

\subsection{Schubert Varieties}
For $W\in\operatorname{Gr}(k,n)$, and intersection threshold $c$, and sample dimension $l$, define  
\[
\Omega_{c,k,l}(W)=\{V\in\operatorname{Gr}(l,n)\mid \dim(V\cap W)\geq c\}.
\]  
This incidence-constrained subset of $\operatorname{Gr}(l,n)$ forms a Schubert variety.

A Grassmannian is itself a single-step flag manifold ($\operatorname{Gr}(k,n)=\operatorname{Fl}(k;n)$). Classical representatives solve mean or median minimization over principal angle distances; SVBF instead solves a maximal incidence approximation.

\section{Algorithm: SVBF–LBG}
We implement a nested optimization: an outer LBG sample-relabeling loop, and an inner Schubert variety fitting loop that learns a subspace $K^*$ minimizing or maximizing first-angle chordal incidence \cite{ourfirst}:

\[
\mathbf{K}^*=\arg \min_{[\mathbf K]\in\operatorname{Gr}(k,n)}
   \sum_{i=1}^p\sin^2\theta_1(\mathbf X_i,\mathbf K).
\]

Following this framework, LBG clustering proceeds in 4 operational stages:

\begin{enumerate}
\item \textbf{Init:} Random or sampled cluster prototypes $\{K_i\}$.
\item \textbf{Labeling:} Assign samples $\mathbf X_j$ to $K_i$ by $d_{ij}=\sin^2\theta_1(K_i,\mathbf X_j)$.
\item \textbf{Prototype Update:} Solve the SVBF problem for each cluster’s assigned group.
\item \textbf{Distortion Check / Loop:} Repeat until distortion flattens or max iterations reached.
\end{enumerate}

\section{Experiments}
The results for all the experiments are reported as distortion versus number of centers plots.
\subsection{Synthetic Subspace Clustering}
\begin{figure}[t!]
  \centering
  \includegraphics[width=0.9\linewidth]{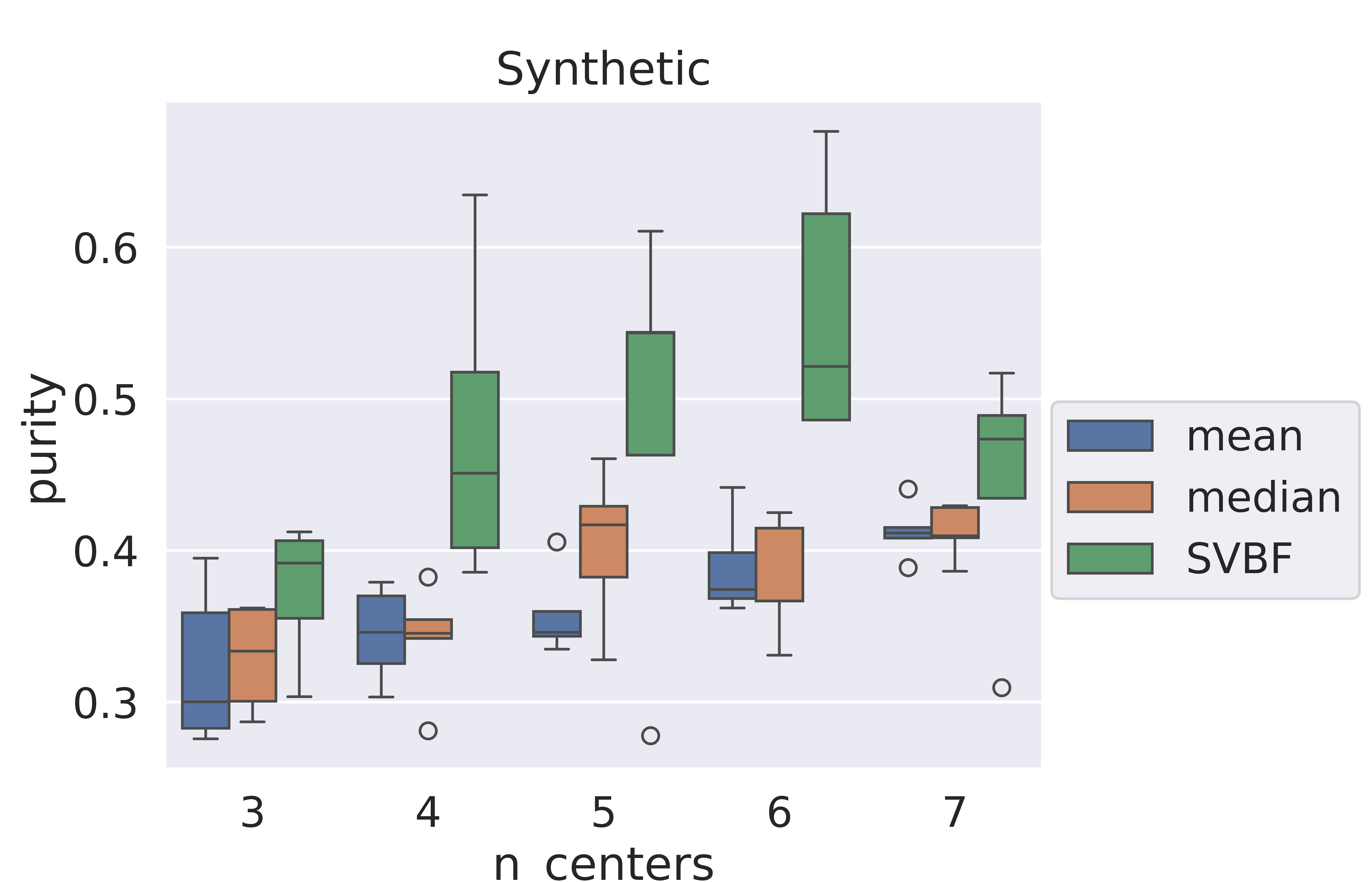}
  \caption{Median purities for synthetic dataset (SVBF, flag mean, flag median).}
  \label{fig:Synthetic}
\end{figure}

We build $50$ ten-dimensional subspaces clustered around $5$ one-dimensional prototypes, then compare LBG-clustering purity across $\{3,4,5,6,7\}$ centers over 5 trials, Figure \ref{fig:Synthetic}.

\subsection{MNIST Subsets \cite{MNIST}}
\begin{figure}[t!]
  \centering
  \includegraphics[width=0.9\linewidth]{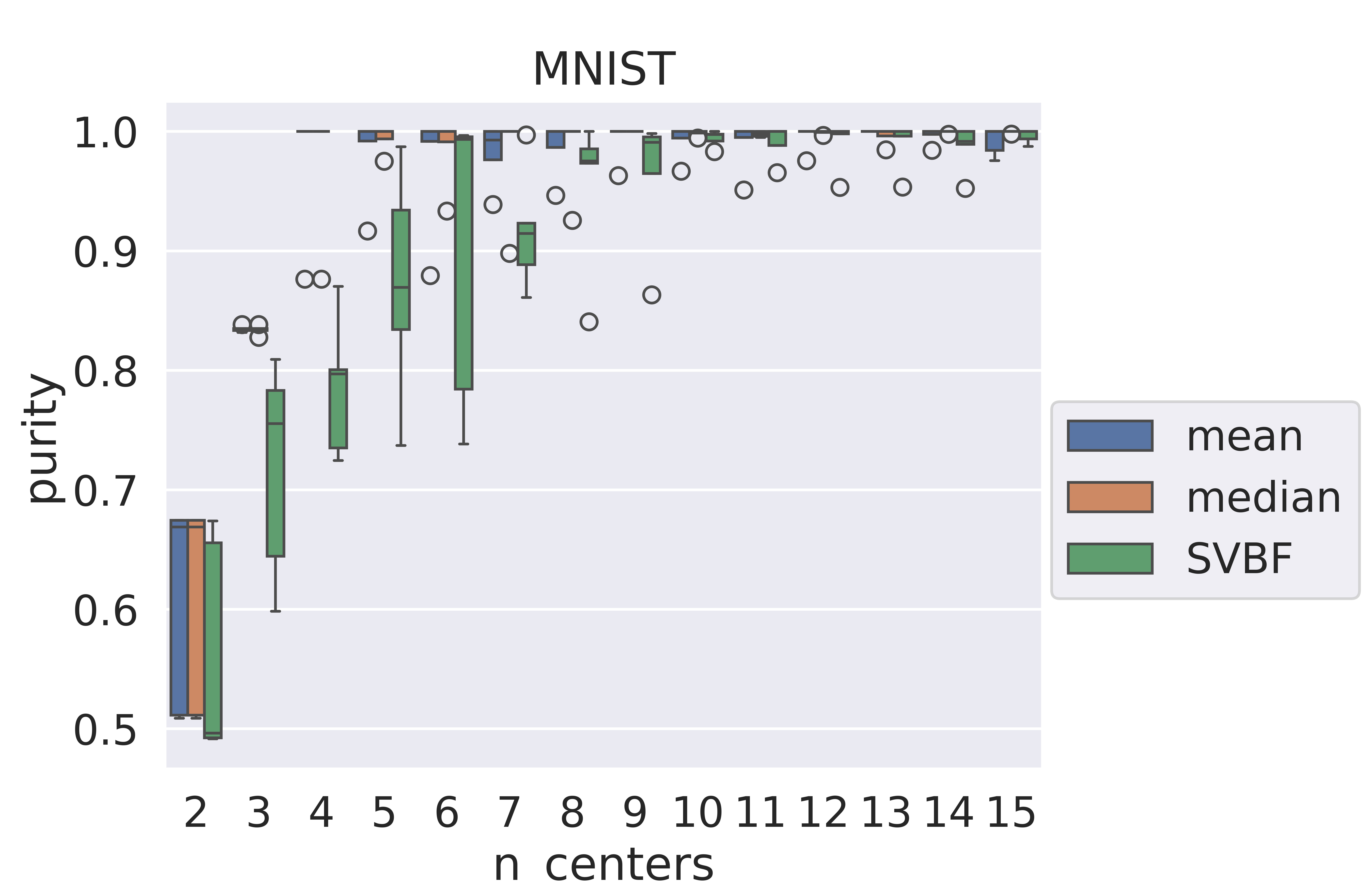}
  \caption{Median cluster purity for MNIST (range 2–15 centers).}
  \label{fig:MNIST}
\end{figure}

Digits $\{0,2,4,6\}$ resized, flattened to $784$-vectors, and grouped into $784\times10$ orthonormal subspace samples. We cluster 5 times with 7 LBG iterations over center range $2$–$15$, Figure \ref{fig:MNIST}.

\subsection{Indian Pines \cite{IndianPines}}
\begin{figure}[t!]
  \centering
  \includegraphics[width=0.9\linewidth]{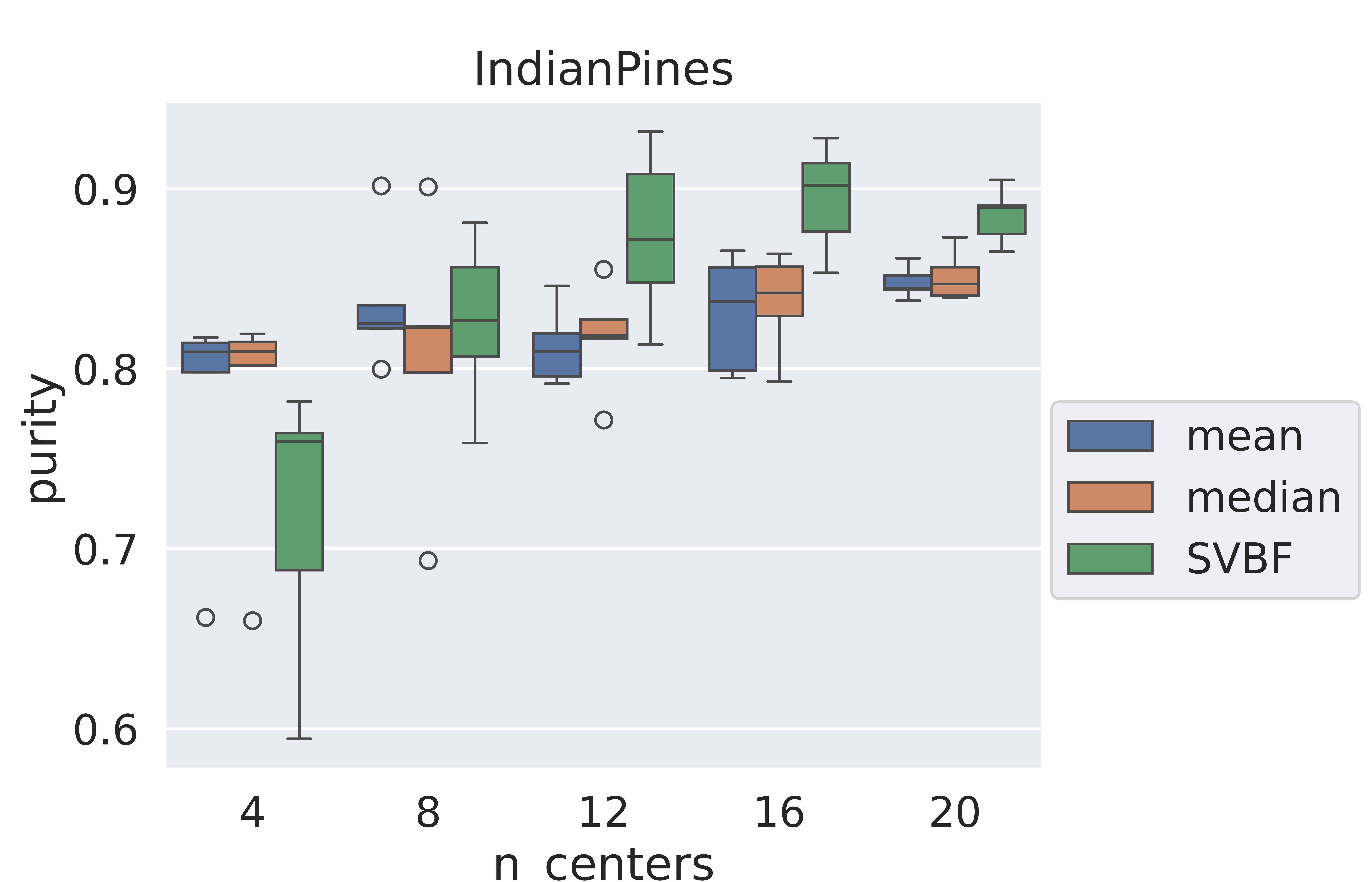}
  \caption{Median cluster purity for Indian Pines.}
  \label{fig:IndianPines}
\end{figure}

200 usable spectral bands, split into $10776$ pixels and largest 4 classes → grouped into $10$-dimensional orthonormal bases, 5 trials with 5 LBG iterations for center set $\{4,8,12,16,20\}$, Figure \ref{fig:IndianPines}.

\subsection{UCF YouTube Action Subset \cite{UCF11}}
\begin{figure}[t!]
  \centering
  \includegraphics[width=0.9\linewidth]{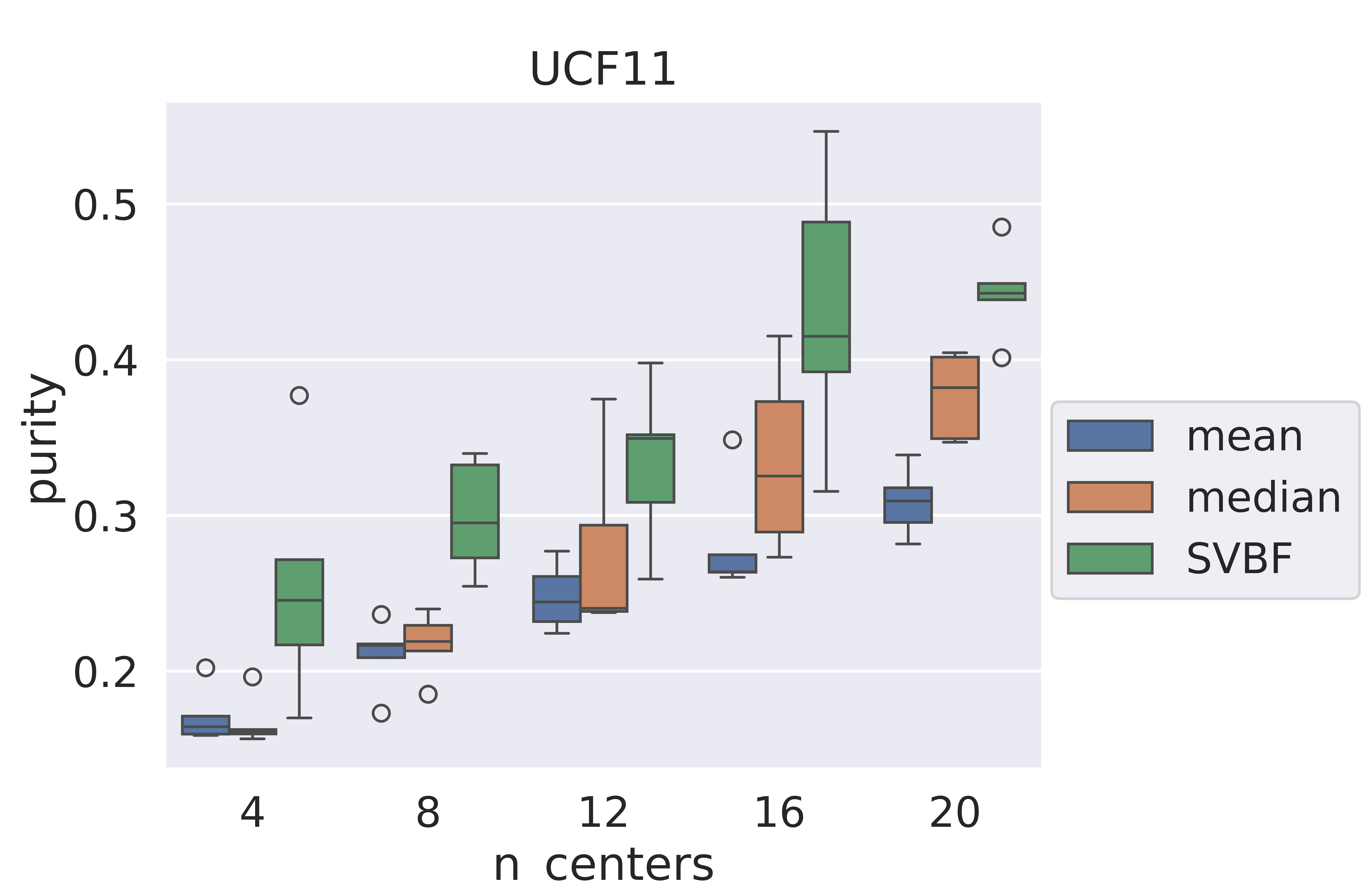}
  \caption{Median cluster purity for UCF YouTube Action.}
  \label{fig:UCF11}
\end{figure}

Video frames → $25\times18$ grayscale → flattened to $450$, grouped as $450\times10$ Grassmann samples, 5 runs, 5 iterations, same center cardinalities, Figure \ref{fig:UCF11}.

\section{Conclusion}
We introduced SVBF as a geometrically flexible trainable prototype integrated into the LBG subspace-clustering pipeline. In incidence-heavy data regimes (e.g., synthetic intersections and hyperspectral geometry), the method surpasses flag mean and flag median cluster purity; in easily separable regimes (small MNIST subsets), all methods converge to near-perfect solutions as center counts grow.  

Current limitations include the fixed dimension of $\mathbf K$ and a non-adaptive stopping rule. GPU acceleration remains required, and future work includes learning multi-directional SVBF incidence constraints and automated convergence analysis. 

\medskip
\noindent{\bf Code availability:}
The code will be available via the link (reach out for getting access now): https://github.com/kkarimov/SubspaceClustering .

\bibliographystyle{elsarticle-num} 
\bibliography{cas-refs}

\end{document}